%% file: root.tex
\documentclass[letterpaper, 10 pt, conference]{ieeeconf}  

\IEEEoverridecommandlockouts                              

\usepackage{graphics} 
\usepackage{epsfig} 
\usepackage{float}
\usepackage[table]{xcolor}
\usepackage{pgfplots}
\usepackage{subcaption} 
\pgfplotsset{compat=newest} 
\usepackage{hyperref}
\hypersetup{colorlinks=true,linkcolor=blue,urlcolor=blue}
\usepackage{amsmath} 
\usepackage{cleveref} 
\usepackage{cite} 

\title{\LARGE \bf
3D Spatial Understanding in MLLMs: Disambiguation and Evaluation
}

\author{Chun-Peng Chang$^{1}$, Alain Pagani$^{1}$ and Didier Stricker$^{1}$ 
\thanks{$^{1}$DFKI Augmented Vision (German Research Center for Artificial Intelligence) 
        {\tt\small \{chun-peng.chang, didier.stricker, alain.pagani\}@dfki.de}}%
}

\begin{document}

\maketitle
\thispagestyle{empty}
\pagestyle{empty}

\begin{abstract}
Multimodal Large Language Models (MLLMs) have made significant progress in tasks such as image captioning and question answering. However, while these models can generate realistic captions, they often struggle with providing precise instructions, particularly when it comes to localizing and disambiguating objects in complex 3D environments. This capability is critical as MLLMs become more integrated with collaborative robotic systems. In scenarios where a target object is surrounded by similar objects (distractors), robots must deliver clear, spatially-aware instructions to guide humans effectively. We refer to this challenge as contextual object localization and disambiguation, which imposes stricter constraints than conventional 3D dense captioning, especially regarding ensuring target exclusivity.
In response, we propose simple yet effective techniques to enhance the model's ability to localize and disambiguate target objects. Our approach not only achieves state-of-the-art performance on conventional metrics that evaluate sentence similarity, but also demonstrates improved 3D spatial understanding through 3D visual grounding model. \\
\url{https://birdy666.github.io/projects/3d_spatial_understanding_in_mllms/}
\end{abstract}


\input{sections/intro}
\input{sections/related_work}
\input{sections/approach}
\input{sections/experiment}

\input{sections/limitations}

\section{CONCLUSIONS}

In this work, we have introduced effective techniques to enhance the spatial reasoning and disambiguation capabilities of MLLMs in complex 3D environments. By focusing on contextual object localization and disambiguation (COLD), our approach not only improves performance on standard datasets like Sr3D and Nr3D but also addresses the limitations of conventional metrics by incorporating 3D visual grounding models. Our results demonstrate that MLLMs, when properly enhanced, can provide clearer, more precise instructions, crucial for real-world applications involving human-robot collaboration in complex 3D spaces. These advancements pave the way for more robust and spatially aware interactions between machines and their environments.

\noindent{\bf Acknowledgement:}
This research has been partially funded by EU project FLUENTLY (GA: Nr 101058680) and ExtremeXP (GA: Nr 101093164).

\addtolength{\textheight}{-12cm}   


\bibliographystyle{splncs04}
\bibliography{egbib}
\end{document}

%% file: sections/intro.tex
\section{INTRODUCTION}

The evolution of multimodal Large Language Models (MLLMs) marks a significant milestone in the field of artificial intelligence, expanding machine understanding and interaction beyond traditional text to include audio \cite{zhang23speechgpt}, video \cite{zhang23videollama}, and images \cite{liu2023llava, alayrac22, li23blip2}. These advancements have enabled LLMs to comprehend and generate content across various modalities. However, the integration of LLMs with 3D point cloud data is relatively new, offering fresh perspectives and capabilities in 3D spatial cognition and multimodal processing.

As system designers begin integrating MLLMs into robotic systems, natural language becomes a more intuitive way to interact with robots. The development of 3D visual grounding (3DVG) has enabled machines to accurately identify objects within complex 3D spaces, marking significant advances in machines' ability to interpret 3D environments~\cite{wang2024uni}. Moving beyond mere understanding, there is now a clear need to extend these capabilities to enable machines to guide human actions within these environments. This is where contextual object localization and disambiguation (COLD) becomes essential. The task addresses the challenge of providing precise and actionable instructions in environments with many similar objects (distractors), thereby enhancing the practical utility of MLLMs in real-world 3D space.

Existing MLLMs which rely on generic Question Answering (QA) datasets \cite{azuma22} to train models, such as 3D-LLM \cite{hong23}, with the expectation that these models can generalize their understanding to a wide range of tasks, including captioning, visual grounding, and QA, within 3D environments. However, when it comes to spatial related instructions, these datasets are often structured around queries like \textit{"Where is the blue chair?"} which presuppose a unique, clearly distinguished object, or "Where is the cup on the nightstand?" where the query inadvertently reveals the object’s exact location. 

\begin{figure}
    \centering
    \subcaptionbox{Hallucination: \\ \textit{"The backpack next to the sink."} \label{fig:test}} [0.48\linewidth]{
        \includegraphics[width=\linewidth]{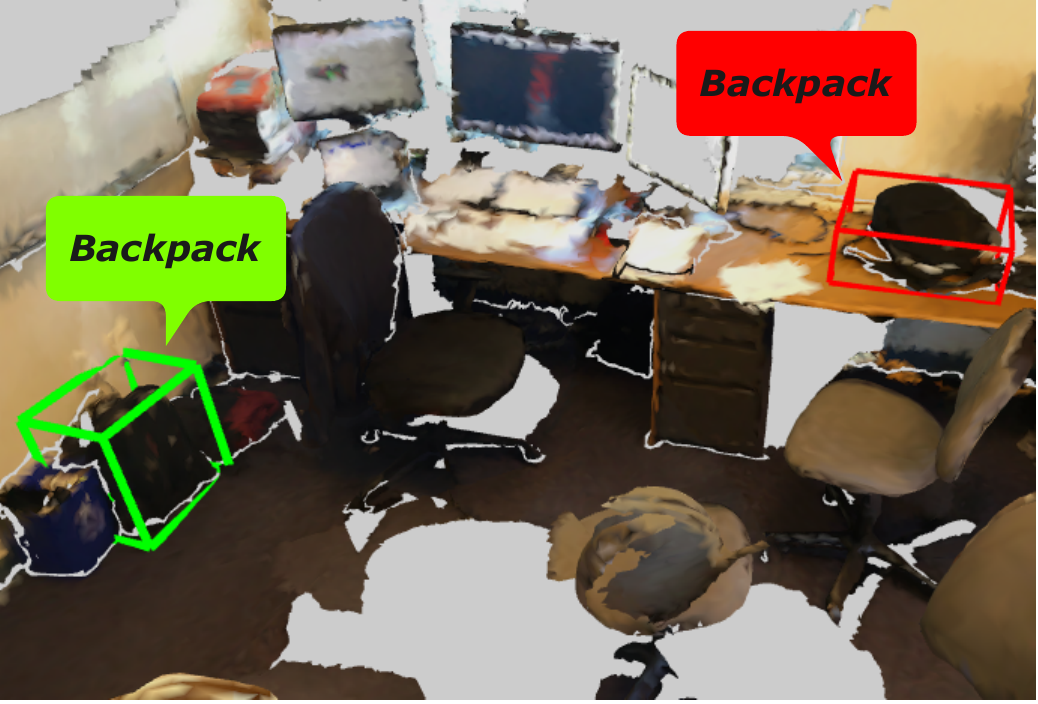}
    }
    \subcaptionbox{Ambiguous Anchor: \\ \textit{"The armchair close to the table."}}[0.48\linewidth]{
        \includegraphics[width=\linewidth]{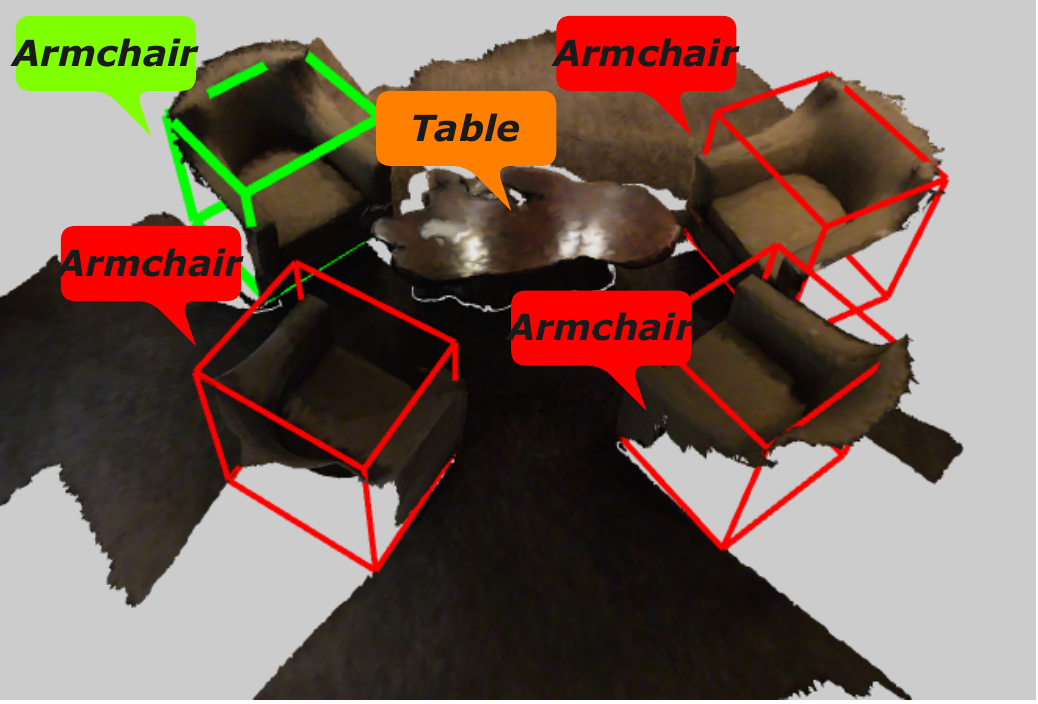}
    }\\
    \subcaptionbox{Wrong Anchor: \\ \textit{"The chair near the blue monitor."}}[0.48\linewidth]{
        \includegraphics[width=\linewidth]{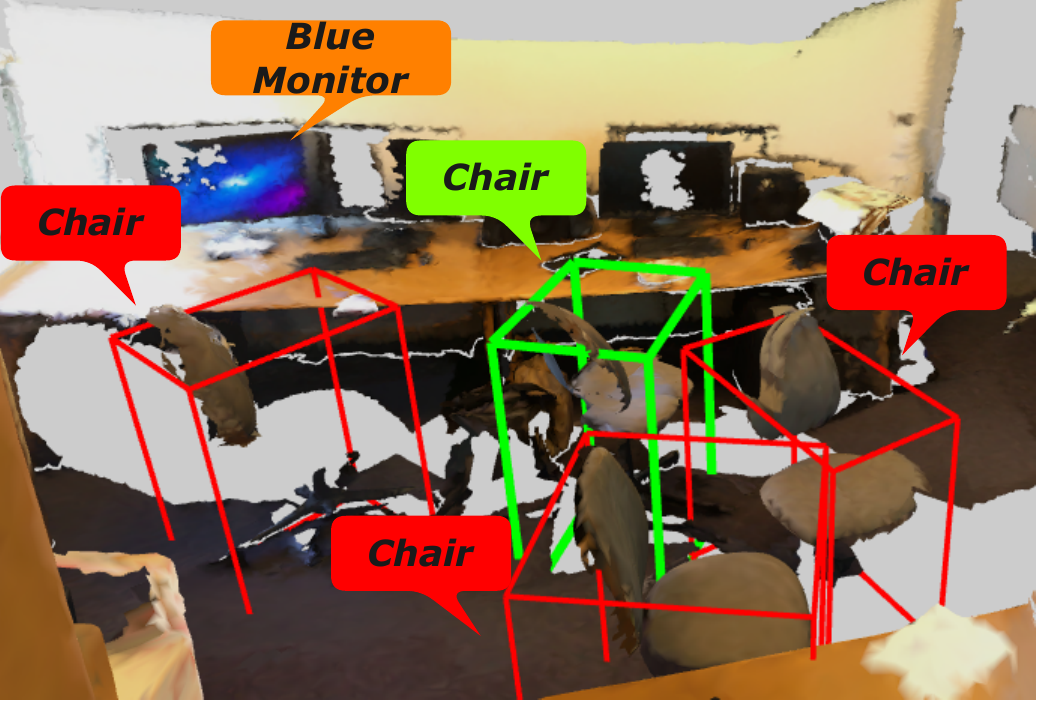}
    }
    \subcaptionbox{Wrong Description: \\ \textit{"The sink far from the trash can."}}[0.48\linewidth]{
        \includegraphics[width=\linewidth]{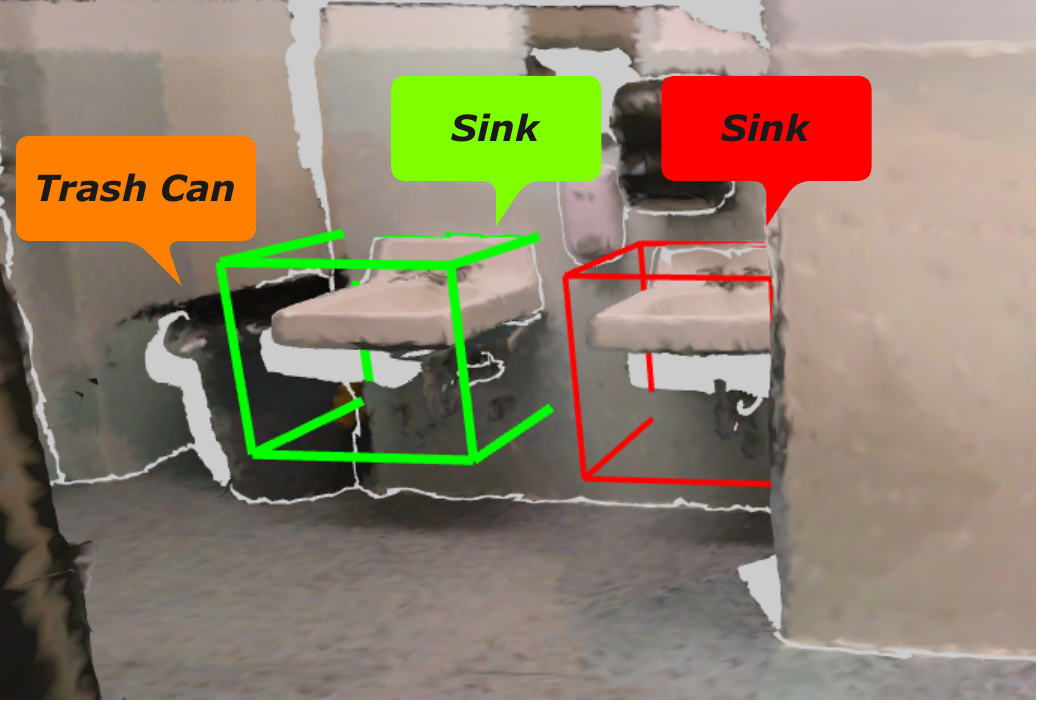}
    }
    \caption{Common mistakes by MLLMs for precise target-exclusive localization and disambiguation.
    (a) depicts a hallucination error where the model refers to an absent object. In (b), the chosen anchor is ambiguous, making localization difficult. (c) illustrates an unsuitable anchor object choice that does not facilitate localize the target. (d) shows a scenario where the model selects an appropriate anchor object but fails to provide a correct spatial description.}
    \label{fig:failed_cases}
\end{figure}

Other works focused on 3D dense captioning \cite{chen2020scan2cap, chen2023end, chen2023vote2capdetr} aim to generate detailed captions for objects within a space, focusing on the spatial attributes or appearance of the objects. These studies attempt to describe each object by its appearance and location. However, a key difference in this task is that disambiguation is not a strict requirement—captions do not need to be exclusive to a single object to be considered correct. A caption that applies to multiple objects may still be valid in captioning tasks. Additionally, while conventional metrics commonly used in 3D dense captioning such as BLEU, ROUGE-L, and CIDEr are useful for evaluating sentence similarity, they are limited in their ability to fully assess a model's 3D spatial understanding. Such formulations often leading to errors like those illustrated in \Cref{fig:failed_cases}. In (a), the model exhibits a hallucination error by referring to an absent object (sink). In (b), the chosen anchor object is ambiguous, complicating localization. In (c), the anchor object choice is unsuitable, failing to help localize the target. Lastly, (d) shows a case where the model selects an appropriate anchor object but fails to provide an accurate spatial description.

In response, we propose simple yet effective techniques that enhance existing LLMs' ability to better understand 3D space and generate more accurate instructions for disambiguating target objects in 3D space. Our proposed methods not only achieve state-of-the-art  performance on the Sr3D and Nr3D datasets, but we also validate the quality of the sentences generated by our model using various 3D visual grounding models.

In conclusion, the main contributions of our work can be summarized as: \\
(1) We enhance the MLLM’s ability for 3D spatial reasoning and precise disambiguation of target objects among similar objects, achieving state-of-the-art performance on the existing datasets.\\
(2) We address the limitation of traditional n-gram-based metrics (e.g., BLEU, CIDEr) and incorporating 3D visual grounding to validate the model's spatial reasoning capabilities. This combined approach offers a more comprehensive and robust evaluation of 3D understanding, ensuring that both linguistic accuracy and spatial comprehension are effectively assessed.

%% file: sections/related_work.tex
\section{RELATED WORK}
\subsection{Multi-Modal LLM}
Advancements in LLMs have significantly propelled the field of multi-modal research, merging language processing with visual, auditory, and spatial data. The LLAVA series\cite{liu2023llava, liu2023improvedllava, liu2024llavanext}, Flamingo\cite{alayrac22}, and BLIP2\cite{li23blip2} have led the charge in image-text reasoning, providing nuanced analyses of visual data in conjunction with textual descriptions.

Furthermore, the exploration of 3D environments through LLMs has introduced innovative models like Chat-3D\cite{wang23} and 3D-LLM\cite{hong23}, which analyze entire scenes through 3D point clouds, enriching the models' spatial awareness. 
\subsection{3D Visual Grounding}
3D visual grounding(3DVG) is a process that involves localizing objects within a 3D space using descriptive text. 
Most research in this field utilizes datasets from Referit3D\cite{achlioptas2020}, which includes Nr3D and Sr3D, as well as ScanRefer\cite{chen2020}.
From an architectural standpoint, there has been a shift from graph-based methods\cite{achlioptas2020, huang2021} to transformer-based methods\cite{abdelreheem2022, zhao2021, yang2021, jain2021, huang2022}. This change reflects a pursuit of higher accuracy in the models.
Addressing the challenge of insufficient data in 3D point clouds, MVT\cite{huang2022} leverages multiple viewpoints to enhance the model's understanding of a scene. ViewRefer\cite{guo23} extends this approach by employing GPT-3\cite{brown20} to augment description capabilities.
MiKASA\cite{chang2024mikasa}, another notable development, introduces a new pipeline, which not only boosts accuracy but also enhances the explainability of the decision-making process in 3D visual grounding tasks.
\subsection{3D Dense Captioning}
Similar to 2D image captioning, the 3D dense captioning task aims to generate text descriptions that characterize specific targets. The term 'dense' signifies that the task involves not merely describing the overall scene but detailing all detected objects within the space through point cloud input, focusing on various aspects such as location and appearance. In Scan2Cap\cite{chen2020scan2cap}, the authors established one of the first benchmarks for this task, which has since seen enhancements in model performance through the adoption of transformer architectures\cite{zhong2022contextualmodeling3ddense, chen2023end, cai20223djcg}. Studies like 3DJCG\cite{cai20223djcg} and D3Net\cite{chen2022d3net} also explored the joint enhancement of 3D dense captioning and visual grounding.

%% file: sections/approach.tex
\section{APPROACH}
\begin{figure*}
  \centering
  \includegraphics[width=\linewidth]{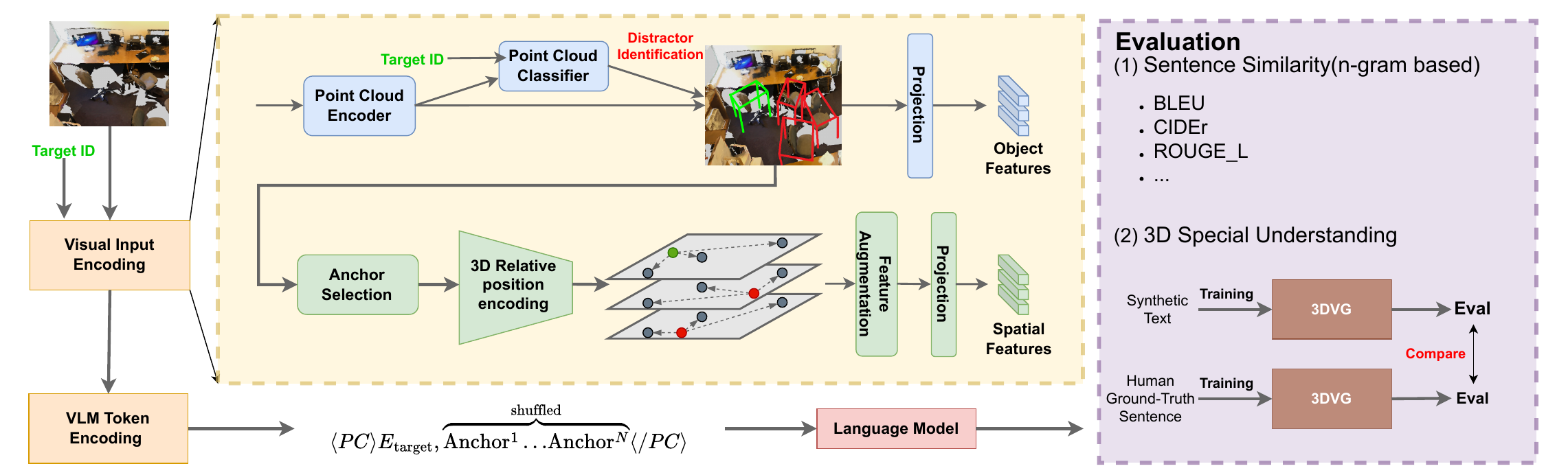}
  \caption{Our training and evaluation pipeline consists of four key steps: visual input encoding, VLM token encoding, LLM generation, and evaluation. The input to the system includes a point cloud and a target ID, which typically comes from upstream tasks such as robot assistants or human-robot teaching interactions, could be a bounding box or just an ID number. In the visual input encoding step, we identify distractors based on the point cloud features of the target object and encode the relative spatial relationships between the target, distractors, and potential anchors. Various token encoding techniques are then applied. During evaluation, we assess the quality of the generated spatial instructions by measuring both sentence similarity and deeper understanding of 3D spatial comprehension.}
  \label{fig:architecture}
\end{figure*}
\begin{table}
\begin{center}
\begin{tabular}{| l| c | c | c|  } 
\hline
\rowcolor[gray]{0.95}
& Length & Open End & Target Exclusive \\
\hline
Translation & - & No & - \\ 
\hline
Caption & $>$ 20 & Yes & No \\
\hline
COLD& $<$ 20 & Yes & Yes \\ 
\hline
\end{tabular}
\end{center}
\caption{Comparison of different text generation tasks, highlighting why the same metrics cannot be applied universally due to the nature of each task. Unlike translation, which typically requires precise equivalency, tasks like captioning and COLD are open-ended, allowing multiple correct interpretations. Moreover, COLD uniquely requires target exclusivity within the 3D space, a criterion not necessary for general captioning.}
\label{table:text_difference}
\end{table}
\subsection{Pipeline}
Our proposed pipeline, illustrated in \Cref{fig:architecture}, consists of a visual input encoding module, a VLM token encoding module, a text generation stage, and an evaluation stage that includes both sentence similarity measurement and 3D spatial understanding assessment. 
Unlike conventional approaches that process all objects in the scene and encode object locations from a single viewpoint, our method explicitly identifies objects similar to the target and enhances spatial understanding by encoding the relative positions between the target object, identified distractors, and potential anchor objects. The encoded object features and spatial relationships are then refined through our token encoding module before being passed into the LLM for text generation.

Evaluation process is also important in pipeline, as the conventional n-gram based methods might not fit all the text genearation task, In \Cref{table:text_difference} we discuss the differences between various text generation tasks and how these differences crucially influence the suitability of conventional metrics. 

Translation, a staple in NLP, involves converting text from one language to another. While there is some flexibility in phrasing, translation does not qualify as an open-ended task because it generally demands precise equivalence between the source and target texts. Unlike translation, captioning and spatial instruction in COLD are open-ended tasks where multiple divergent responses can all be considered correct. 

However, a critical distinction exists between these tasks: Captioning often involves describing a scene or object broadly, typically with longer text, without the need for pinpoint accuracy or exclusivity. In contrast, our goal in COLD requires descriptions to be exclusively linked to the target object and tends to be more concise. This demand for object-specific descriptions adds a layer of complexity to the task.
The length of the generated text plays a crucial role in the effectiveness of evaluation metrics. For longer sentences, higher n-gram based metric scores can provide more confidence that the reference sentence and the one generated by the MLLM are similar in context but this confidence diminishes with shorter, concise instructions where the metrics may not adequately capture the intended meaning or context. 

To address this, we use the text generated by the MLLM as synthetic training data to train various 3DVG models. Comparisons with human-created ground truth data show that our model generates textual instructions that not only closely resemble human-made text but also enable 3DVG models to learn spatial understanding in a human-like manner. While we acknowledge that this evaluation could be influenced by the imperfections of the current 3DVG models, we believe that considering both conventional metrics and 3DVG models provides a more comprehensive analysis. Furthermore, we anticipate that as 3DVG models continue to improve, the evaluation process will become even more reliable and promising in the near future.

\subsection{Explicit Distractors Identification:}
To generate a sentence that exclusively describes a target object in a complex 3D space, it is crucial to identify all potential distractors to ensure the sentence does not apply to multiple objects. While large language models may develop some capacity for this through extensive training on large datasets, our experiments show that explicitly identifying distractors significantly improves the quality of the generated descriptions. Given an encoded target object, we use a pretrained classifier to identify all similar objects in the scene. In the next step, we encode the relative position features based on both target/distractors and potential anchor objects. As shown in \Cref{eq:distractor_selection}, \(D\) denotes the set of distractors, \(o_i\) represents the \(i\)-th object in the scene, and \(c_{\text{target}}\) is the category of the target object. An object \(o_i\) is flagged as a distractor if the category with the highest score from the classifier matches the target object's category.

\begin{equation}
D = \{o_i | \arg\max_{k}(\mathcal{C}(o_i)[k]) = \mathcal{C}(o_{target})\}
\label{eq:distractor_selection}
\end{equation}

\subsection{Relative Position Encoding}

Inspired by contemporary advancements in 3D visual grounding models\cite{chang2024mikasa}, we propose that utilizing relative location encoding significantly enhances a model's comprehension of the 3D world. Traditional methods often rely on absolute location data from datasets, which may not effectively convey the nuanced spatial relationships between objects within a scene. By shifting our focus to explicitly encode the relative spatial features, we provide a more robust framework for the model to grasp and interpret the spatial dynamics among objects. In order to avoid the significant increase of complexity when there are a lot of object in the space, we only encode the features between target object, distractors and the potential anchors.

Given the absolute position \((x_i, y_i, z_i)\) of object \(o_i\) and the absolute position \((x_j, y_j, z_j)\) of any other object \(o_j\) in the scene, the relative position of \(o_j\) with respect to \(o_i\) is computed as shown in \Cref{eq:relative_position}. This includes a normalization function, \(\mathcal{N}(\cdot)\), applied to ensure consistency in scale and spatial relationships. The normalized relative coordinates of \(o_j\) with respect to \(O_i\) are denoted as \((x^i_j, y^i_j, z^i_j)\). We apply this operation to all \(n\) objects in the space, generating a relative position map of size \(n \times n \times 3\).

\begin{equation}
(x^i_j, y^i_j, z^i_j) = \mathcal{N}(x_j - x_i, y_j - y_i, z_j - z_i)
\label{eq:relative_position}
\end{equation}

\subsection{Random Anchors and Ambiguous Anchors:}
To enhance the model's learning on selecting appropriate anchors, we randomly add ambiguous anchor object proposals during training stage based on their spatial relationship to the target during training stage. By choosing objects whose proximity to the target is neither the closest nor the farthest, we aim to increase the likelihood of including ambiguous anchors within the model's training regime.

\subsection{Input Preprocessing and Instruction Tuning:}
To ensure our model does not learn to memorize the position of the ground truth anchor, we shuffle the sequence of tokens for different potential anchor objects during the training stage. For instance, information related to each anchor is encapsulated as \(<Anchor_i>\)...\(</Anchor_i>\). This method randomizes the order of the anchors for each instance, forcing the model to focus on the content and spatial context provided within the tags, rather than their sequential position. 
As shown in \Cref{eq:shuffle}.
{
\small
\begin{equation}
\langle PC \rangle E_{\text{target}} , \overbrace{\text{Anchor$^1 \ldots $Anchor$^N$}}^{\text{shuffled}} \langle / PC \rangle
\label{eq:shuffle}
\end{equation}
}
\\
where each Anchor$^i$ is further defined as \(\langle Anchor^i \rangle E_{\text{anchor}}^i , E_{\text{RP}}^i \langle /Anchor^i \rangle\). Here, \(E_{\text{target}}\) stands for the embedding of the target object, \(Emb_{\text{anchor}}^i\) for the embedding of anchor \(i\), and \(E_{RP}^i\) for the relative position embedding of anchor \(i\). 

\subsection{Loss Function}
\label{sections_sup:loss_calculation}
\subsubsection{Stage 1}
The object-text alignment is refined through a loss function that combines Mean Squared Error (MSE) and cosine similarity. See \Cref{eq:loss_stage1}, \( \mathbf{v}_i \) represents the projected feature vector for the i-th example, \( \hat{\mathbf{v}}_i \) is the target text embedding from the LLM, and \( N \) is the number of examples in the batch. The hyperparameters \( \alpha \) and \( \beta \) are fine-tuned based on grid search.
\begin{equation}
\mathcal{L}_{1} = \alpha \left( \frac{1}{N} \sum_{i=1}^{N} (\mathbf{v}_i - \hat{\mathbf{v}}_i)^2 \right) + \beta \left( 1 - \frac{\mathbf{v}_i \cdot \hat{\mathbf{v}}_i}{\|\mathbf{v}_i\| \|\hat{\mathbf{v}}_i\|} \right)
\label{eq:loss_stage1}
\end{equation}

\subsubsection{Stage 2}
The fine-tuning of the MLLM apply the standard cross-entropy loss to guide the model's language generation process, as shown in \Cref{eq:loss_stage2}.  

\begin{equation}
\mathcal{L}_{2} = -\sum_{i} y_i \log(\hat{y}_i)
\label{eq:loss_stage2}
\end{equation}
Where $y_i$ is the ground truth distribution and $\hat{y}_i$ is the model's predicted probability distribution over the vocabulary.

%% file: sections/experiment.tex

\section{EXPERIMENTS}
\begin{table}
\centering
\begin{tabular}{c | c | c | c | c } 
\hline
\rowcolor[gray]{0.95}
& PC encoder & Proj(vision) & Proj(spatial) &  LoRA\cite{hu22} \\
\hline
\#param & 13M & 11M & 10M & 81M \\ [0.5ex]
\hline
Update & - & Stage 1 & Stage 2 & Stage 2 \\
\hline
\end{tabular}
\caption{Parameter count and update schedule for each module. The table's second row indicates the update timing for each module, providing insights into the model's adaptation and refinement process.}
\label{table:params}
\end{table}

\subsection{Implementation Details}
\label{sections_sup:implementation_details}
For point cloud encoding, we utilized a pretrained visual module that combines PointNet++\cite{qi2017a} with a transformer encoder\cite{vaswani2017}. Language processing was managed using the pretrained LLAMA\cite{touvron23} (7B) model, further enhanced by integrating Vicuna(7B)\cite{chiang23} delta weights to tailor its performance for our specific requirements. Our computational framework was supported by 1 NVIDIA A100 GPUs. 
\Cref{table:params} details the number of parameters for each module. In Stage 1, where object features are aligned with textual features, we employed a multi-layer perceptron (MLP) with 11M parameters. In Stage 2, we fine-tuned the Vicuna 7B model with LoRA\cite{hu22}, involving an additional 81M parameters, and updated the projection layer of the spatial module, adding 10M parameters. Consequently, a total of 91M parameters were updated in this stage.
\subsection{Efficiency of Our Proposed Pipeline}
\label{sec:ablation}
\begin{table*}
\centering
\begin{tabular}{|c|l|cccccc|} 
\hline
\rowcolor[gray]{0.95}
Dataset & \multicolumn{1}{c|}{Method} & B-1 & B-2 & B-3 & B-4 & CIDEr & ROUGE-L \\
\hline
& Vote2Cap\cite{chen2023end} & 0.40 & 0.28 & 0.18 & 0.12  & 0.27 & 0.30\\ 
\cline{2-8}
Nr3D & Vote2Cap++\cite{chen2023vote2capdetr} & 0.41 & 0.32 & 0.24 & \bf{0.18}  & \bf{0.29} & 0.32\\ 
\cline{2-8}
& Ours & \bf{0.57}  & \bf{0.38} & \bf{0.24} & 0.15 & 0.26 & \bf{0.43} \\ 
\hline
& Vote2Cap\cite{chen2023end} & 0.49 & 0.45 & 0.41 & 0.38  & 2.24 & 0.46\\ 
\cline{2-8}
& Vote2Cap++\cite{chen2023vote2capdetr} & 0.52 & 0.49 & 0.45 & 0.42  & 2.47 & 0.49 \\ 
\cline{2-8}
Sr3D& Ours & \bf{0.60}  & \bf{0.54} & \bf{0.49} & \bf{0.44} & \bf{2.54} & \bf{0.57} \\ 
\cline{2-8}
& Ours + "far" &0.60  & 0.52 & 0.45 & 0.38   & 2.07 & 0.56 \\ 
\cline{2-8}
&  Ours + "close" & 0.58  & 0.49 & 0.39 & 0.31   & 1.59 & 0.54 \\ 
\hline

\end{tabular}
\caption{Performance comparison of different models on the Nr3D and Sr3D test datasets, showing that our approach outperforms Vote2Cap++\cite{chen2023vote2capdetr} across several metrics.
However, we also observe that while these conventional metrics effectively measure sentence similarity, they fail to capture the nuanced 3D spatial understanding. In the Sr3D evaluation, we manually replaced all words related to spatial relationships with either \textit{"far"} or \textit{"close"}, resulting in only a slight drop in performance. This suggests that n-gram-based evaluations alone may not be sufficient to fully assess a model's ability to generate precise instructions that can disambiguate target objects from distractors.}
\label{table:convention}
\end{table*}

\begin{figure*}
  \centering
  \includegraphics[width=\linewidth]{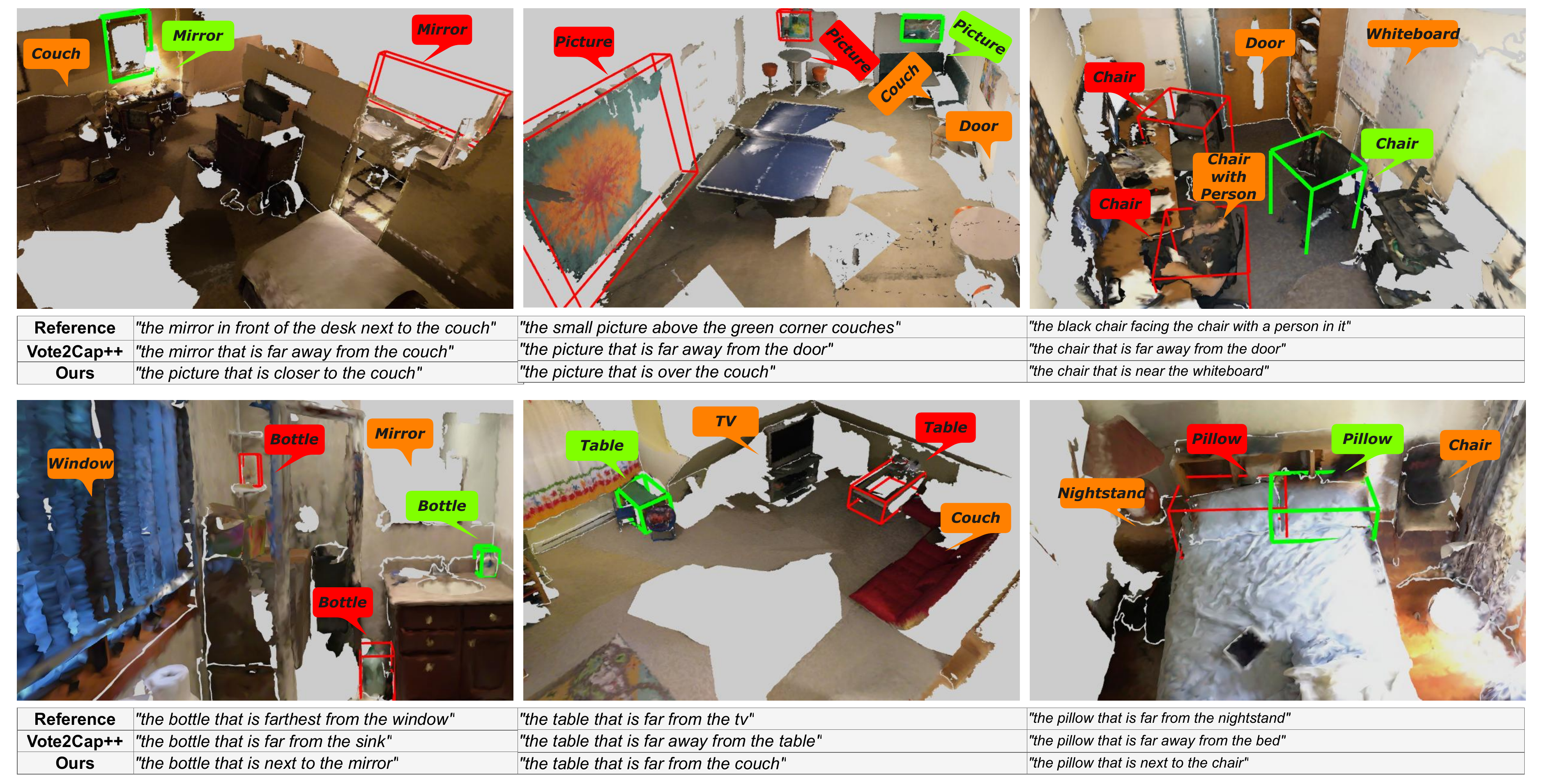}
  \caption{Comparison of captions generated by different language models. Both Vote2Cap++ and our model generate sentences with structures highly similar to the reference sentence. However, the key difference lies in the spatial understanding: the sentences generated by our proposed method demonstrate a better grasp of the spatial configuration within the scene, even when a different anchor is chosen. This shows that our approach helps the model develop a deeper understanding of the 3D scene, rather than merely mimicking the surface-level sentence structure.}
  \vspace{-3mm}
  \label{fig:visresult}
\end{figure*}

In \Cref{table:convention} and \Cref{fig:visresult} we compare our approach with the state-of-the-art 3D dense captioning model for two key reasons. First, this comparison allows us to validate our model's performance using conventional metrics that measure sentence similarity, providing insight into how well our approach aligns with textual benchmarks. Second, it emphasizes the uniqueness of the COLD task, which requires a deeper spatial understanding.

To thoroughly assess the quality of the generated text, particularly in complex spatial reasoning tasks, we evaluated our results using the Nr3D and Sr3D datasets~\cite{achlioptas2020}, where ground truth object bounding boxes are provided. We trained both models and compared their performance on the test datasets using conventional metrics. The results show that our model outperforms Vote2Cap++~\cite{chen2023vote2capdetr} across several metrics, indicating that our generated text is, at the very least, similar to the ground truth at a surface level.

Moving beyond surface-level similarity, we noticed that while the language model can generate sentences with a structure similar to the reference sentence, the key spatial relationship terms are sometimes not accurate. To further investigate this phenomenon, we manually replaced all spatial relationship terms in our generated sentences with either \textit{"far"} or \textit{"close"}. While the scores on conventional metrics dropped, the decrease was minimal, suggesting that while these metrics provide valuable insights into shallow text similarity, they are insufficient for fully evaluating the quality and precision of the generated text in capturing spatial relationships.
It is important to note that we are not dismissing the value of conventional metrics, as they do provide useful insights into text similarity. However, additional evaluation methods are needed to fully assess the quality and accuracy of the generated text, particularly in tasks involving complex spatial reasoning.

\definecolor{darkgray}{rgb}{0.5, 0.5, 0.5}
\begin{table}
\begin{center}
\begin{tabular}{|l|l | llll| } 
\hline
\rowcolor[gray]{0.95}
 3DVG Model & Train Data & Easy & Hard & VD\textcolor{red}{\textsuperscript{$\ast$}} & VI\textcolor{blue}{\textsuperscript{$\dagger$}}  \\ [0.5ex]
\hline
& \textcolor{darkgray}{Sr3D} & \textcolor{darkgray}{38.6\%} & \textcolor{darkgray}{28.9\%} & \textcolor{darkgray}{36.8\%} & \textcolor{darkgray}{35.6\%} \\ 
\cline{2-6} 
ReferIt3D\cite{achlioptas2020} & Vote2Cap++  & 22.3\% & 17.4\% & 14.7\% & 20.9\% \\
&Ours  & {\bf 29.4\%} & {\bf 23.3\%} & {\bf 15.8\%} & {\bf 28.1\%} \\
\hline
& \textcolor{darkgray}{Sr3D} & \textcolor{darkgray}{56.0\%} & \textcolor{darkgray}{50.3\%} & \textcolor{darkgray}{44.8\%} & \textcolor{darkgray}{54.9\%} \\ 

\cline{2-6} 
MVT\cite{huang2022}&Vote2Cap++ & 24.6\% & 21.9\% & 16.3\% & 24.1\% \\
&Ours & {\bf 39.8\%} & {\bf 35.2\%} & {\bf 20.5\%} & {\bf 39.2\%} \\

\hline
& \textcolor{darkgray}{Sr3D} & \textcolor{darkgray}{63.8\%} & \textcolor{darkgray}{58.4\%} & \textcolor{darkgray}{60.2\%} & \textcolor{darkgray}{62.2\%} \\ 

\cline{2-6} 
MiKASA\cite{chang2024mikasa}&Vote2Cap++ & 31.5\% & 23.8\% & 17.4\% & 29.6\% \\
&Ours & {\bf 48.7\%} & {\bf 41.0\%} & {\bf 18.3\%} & {\bf 47.7\%} \\

\hline
\end{tabular}
\end{center}
\caption{Comparative Performance of various models on the Sr3D\cite{achlioptas2020} Test Dataset. This table illustrates the evaluation results for models trained on human made dataset(Nr3D/Sr3d) , with instructions generated by different approaches. The training set scenes were divided, allocating a 50-50 portion to prevent any overlap between the scenes used for training the models and those employed in generating instructions.\textcolor{red}{\textsuperscript{$\ast$}}\footnotesize{View-Dependent}, \textcolor{blue}{\textsuperscript{$\dagger$}}\footnotesize{View-Independent}}
\label{table:sr3d5050}
\end{table}
\definecolor{mydarkred}{RGB}{200,0,0}
\begin{table}
\begin{center}
\begin{tabular}{|l|ccc|} 
\hline
\rowcolor[gray]{0.95}
Train Data & \(|D| = 1\) & \(|D| = 2\) & \(|D| = 4\) \\ [0.5ex]
\hline
\textcolor{darkgray}{Nr3D} & 
\textcolor{darkgray}{\(60.5\%\)} & \textcolor{darkgray}{51.2\%\color{mydarkred}$\downarrow$ 15\%} & 
\textcolor{darkgray}{47.0\% \color{mydarkred}$\downarrow$ 22\%} \\ 

\textcolor{darkgray}{Sr3D} & \textcolor{darkgray}{\(78.6\%\)} & 
\textcolor{darkgray}{68.7\%\color{mydarkred}$\downarrow$ 13\%} & \textcolor{darkgray}{59.0\%\color{mydarkred}$\downarrow$ 25\%} \\ 
\hline
Vote2Cap++ & \(42.6\%\) & 
32.5\%\color{mydarkred}$\downarrow$ 24\% & 21.8\%\color{mydarkred}$\downarrow$ 49\% \\ 

Ours&\(60.2\%\) & 
49.6\%{\bf \color{mydarkred}$\downarrow$ 17\%} & 41.8\%{\bf \color{mydarkred}$\downarrow$ 30\%} \\ 
\hline
\end{tabular}
\end{center}
\caption{The table presents a comparative analysis of MiKASA's\cite{chang2024mikasa} performance in scenarios with varying numbers of distractors present. The experiments were conducted using Sr3D dataset. }
\label{table:distractor}
\end{table}

\subsection{3DVG for Spatial Understanding Evaluation }
Recognizing the limitations of conventional metrics, we further evaluated the spatial reasoning ability of our model using 3D Visual Grounding (3DVG) models. In \Cref{table:sr3d5050} we trained 3DVG models using synthetic datasets generated by both our model and Vote2Cap++~\cite{chen2023vote2capdetr}. The rationale behind this is that if the text generated by the language models is realistic and precise enough, the 3DVG models should be able to learn from the synthetic text and perform similarly to those trained on human-made ground truth. The results suggest that the synthetic data generated by our model better trains the 3DVG model, leading to improved performance on the test data compared to Vote2Cap++~\cite{chen2023vote2capdetr}.

\subsection{Performance as the Number of Distractors Increases} Another important evaluation metric for assessing the quality of generated sentences is how the model's performance declines as the number of distractors increases, making the grounding task more challenging. In \Cref{table:distractor}, we observe that when 3DVG models are trained on human-created datasets, performance drops by around 13-15\% when the number of distractors increases to 2, and by 22-25\% when the number of distractors increases to 4.

In comparison, models trained on Vote2Cap++~\cite{chen2023vote2capdetr} synthetic data show a significantly larger performance drop, with a 24\% decrease when there are 2 distractors and a 49\% decrease with 4 distractors. On the other hand, models trained on our synthetic data show a more modest decline, with a 17\% drop for 2 distractors and a 30\% drop for 4 distractors. This performance is much closer to that of models trained on human-created data, demonstrating the superior quality of our synthetic data compared to Vote2Cap++.

\subsection{Ablation Study}
\begin{table*}
\centering
\begin{tabular}{|l|ccc|ccc| }
\hline
\rowcolor[gray]{0.95}
Variant & Referit3D & MVT & MiKASA & B-4 & CIDEr & ROUGE-L\\ \hline
Our Approach & \bf 30.6\%\(\pm0.2\%\) & \bf 51.0\%\(\pm0.1\%\) & \bf 56.8\%\(\pm0.2\%\) & 0.44 & 2.54 & 0.57  \\
w/o Explicit Relative Position Encoding & \(28.8\%\pm0.5\%\) & \(37.4\%\pm0.2\%\) & \(41.6\%\pm0.2\%\) & 0.42 & 2.22 & 0.51 \\
w/o Tokens Shuffling & \(19.7\%\pm0.2\%\) & \(45.4\%\pm0.1\%\) & \(53.0\%\pm0.1\%\)& 0.43 & 2.36 & 0.55\\
w/o Explicit Distractor identification & \(29.9\%\pm0.3\%\) & \(39.9\%\pm0.2\%\) & \(43.0\%\pm0.1\%\) & 0.40 & 2.20 & 0.50\\
w/o Adding Ambiguous Anchors & \(28.4\%\pm0.3\%\) & \(48.9\%\pm0.2\%\) & \(52.0\%\pm0.1\%\) & 0.44 & 2.44 & 0.56\\
\hline
\end{tabular}
\caption{Ablation studies demonstrating the efficiency of individual modules within our framework, the MLLM is trainined with the whole Sr3D training scenes. Performance metrics are gauged using the Sr3D test dataset, with each reported value representing the overall accuracy across all tests. }
\label{tab:ablation_study}
\end{table*}
In \Cref{tab:ablation_study} we demonstrate the effectiveness of our proposed modules and fine-tuning techniques. The results show that relative position encoding significantly enhances the language model's ability to generate spatially-aware sentences, as does the explicit identification of distractors. Additionally, fine-tuning techniques, such as shuffling and introducing ambiguity, lead to a slight improvement in overall sentence quality, further validating the impact of these approaches.

%% file: sections/limitations.tex
\section{Limitations}
\label{section_sup:limitations}
Despite the advancement of exiting works, there are still significant challenges that hinder performance. 
The first notable limitation is the loss of directional information in instructions generated by MLLMs. This shortfall is especially evident in tasks that require a change in the viewer's perspective. The limitation arises from the point cloud encoding process. While point cloud encoders like PointNet++\cite{qi2017a} offer rotation-invariant encoding, which aids in object recognition, the trade-off is the loss of directional cues. This impacts the model's ability to provide effective guidance in scenarios where orientation changes are crucial.
A second limitation concerns the model's performance with non-rigid instances, such as humans and animals, where available training data is sparse. The current architecture and training datasets do not adequately support learning the complex and varied representations of these entities, leading to challenges in accurately identifying and interacting with non-rigid objects within scenes. 
The third limitation emerges when the model trying to generate action-oriented instructions. These situations demand not only an understanding of object orientation but also a nuanced comprehension of how humans interact with objects. In the sentence like \textit{"The chair with a person sitting on it"}, the model must grasp the implied action (sitting) and its relation to the specified object (chair), a task that extends beyond spatial recognition to involve behavioral inference.